%% file: main.tex
\documentclass[letterpaper, 10 pt, conference]{ieeeconf}  % Comment this line out if you need a4paper

\usepackage[nocompress]{cite}

\IEEEoverridecommandlockouts                              % This command is only needed if 
\usepackage{xcolor}
\usepackage{xspace}
\usepackage{dirtytalk}
\usepackage{graphicx}
\usepackage{subcaption}
\usepackage{amsmath}
\usepackage{booktabs}
\usepackage{multirow}
\usepackage{amsfonts}
\usepackage{array}
\usepackage{makecell}
\usepackage{xcolor}
\usepackage{arydshln}
\usepackage{balance}
\newcommand{\NoStateModels}{State-free Policies\xspace}
\newcommand{\YesStateModels}{state-based policies\xspace}
\newcommand{\YesStateModel}{state-based policy\xspace}
\newcommand{\pickplace}{\say{Pick $\&$ Place}\xspace}
\newcommand{\pizero}{$\pi_0$\xspace}
\newcommand{\eefcam}{wrist-camera\xspace}
\newcommand{\eefcams}{wrist-cameras\xspace}

\makeatletter
\let\NAT@parse\undefined
\makeatother
\usepackage{hyperref} 
\hypersetup{
    colorlinks=true,
    linkcolor=red,
    citecolor=green
}

\title{\LARGE \bf
Do You Need Proprioceptive States in Visuomotor Policies?
}

\author{Juntu Zhao$^{1,2*}$, Wenbo Lu$^{2,4*}$, Di Zhang$^{2,5}$, Yufeng Liu$^{1,2}$, Yushen Liang$^{4}$, Tianluo Zhang$^{4}$, Yifeng Cao$^{2}$, \\ Junyuan Xie$^{2}$, Yingdong Hu$^{2,3}$, Shengjie Wang$^{4}$, Junliang Guo$^{2\ddag}$, Dequan Wang$^{1\dag}$ and Yang Gao$^{2,3\dag}$% <-this % stops a space
\thanks{$^{*}$Equal Contribution. This work was done during the internship at Spirit AI. $^{\dag}$Corresponding authors. $^{\ddag}$Project leader.}
\thanks{$^{1}$Shanghai Jiao Tong University, $^{2}$Spirit AI, $^{3}$Tsinghua University, $^{4}$New York University Shanghai, $^{5}$Tongji University.}
}

\begin{document}

\maketitle
\thispagestyle{empty}
\pagestyle{empty}

\input{Sections/0_abstract}
\input{Sections/1_introduction}
\input{Sections/2_related_works}

\input{Sections/3_no_state}

\input{Sections/4_experiments}

\input{Sections/5_conclusion}

\input{Sections/9_appendix}

% \section*{ACKNOWLEDGMENT}

% The preferred spelling of the word ÒacknowledgmentÓ in America is without an ÒeÓ after the ÒgÓ. Avoid the stilted expression, ÒOne of us (R. B. G.) thanks . . .Ó  Instead, try ÒR. B. G. thanksÓ. Put sponsor acknowledgments in the unnumbered footnote on the first page.

%%%%%%%%%%%%%%%%%%%%%%%%%%%%%%%%%%%%%%%%%%%%%%%%%%%%%%%%%%%%%%%%%%%%%%%%%%%%%%%%

% References are important to the reader; therefore, each citation must be complete and correct. If at all possible, references should be commonly available publications.

\balance
% \bibliographystyle{IEEEtran}
% \bibliography{IEEEabrv,main}

\input{main.bbl}
\end{document}

%% file: Sections/0_abstract.tex
\begin{abstract}
% [Version3]
Imitation-learning-based visuomotor policies have been widely used in robot manipulation, where both visual observations and proprioceptive states are typically adopted together for precise control. However, in this study, we find that this common practice makes the policy overly reliant on the proprioceptive state input, which causes overfitting to the training trajectories and results in poor spatial generalization. On the contrary, we propose the State-free Policy, removing the proprioceptive state input and predicting actions only conditioned on visual observations. The State-free Policy is built in the relative end-effector action space, and should ensure the full task-relevant visual observations, here provided by dual wide-angle wrist cameras. Empirical results demonstrate that the State-free policy achieves significantly stronger spatial generalization than the state-based policy: in real-world tasks such as pick-and-place, challenging shirt-folding, and complex whole-body manipulation, spanning multiple robot embodiments, the average success rate improves from 0\% to 85\% in height generalization and from 6\% to 64\% in horizontal generalization. Furthermore, they also show advantages in data efficiency and cross-embodiment adaptation, enhancing their practicality for real-world deployment.

\end{abstract}

%% file: Sections/1_introduction.tex
\section{Introduction}
Imitation-learning-based visuomotor policies~\cite{chi2023diffusion, zhao2023learning, black2410pi0,barreiros2025careful,Zhao2025CoTVLAVC} have been widely used in robotic manipulation. 
Leveraging large-scale demonstration datasets~\cite{o2024open,khazatsky2024droid,walke2023bridgedata,Jones2025BeyondSF} and fine-tuning powerful pre-trained policies have enabled robots to achieve remarkable performance across diverse real-world tasks.

For precise and reliable control, these visuomotor policies typically incorporate not only visual observations of the task environment but also proprioceptive state (hereafter referred to as \textit{state}) inputs~\cite{torabi2019imitation, black2410pi0, kuang2025adapt}, such as end-effector poses and joint angles. The state inputs provide compact and accurate information about the robot’s configuration, but they also make the policy prone to overfitting by simply memorizing the training trajectories.
Therefore it severely limits spatial generalization~\cite{chen2025improving, zeng2021transporter, yin2023spatial} if the training data lacks diversity~\cite{Lin2024DataSL}.
In today’s context, where collecting demonstration data with wide state coverage (i.e., diverse spatial locations of task-relevant objects) is prohibitively expensive, this has become a critical bottleneck for the development of visuomotor policies.

\input{Figures/1_teaser/teaser}

In this study, we propose to completely remove the state input in visuomotor policies to enhance their spatial generalization ability, hereafter referred to as \say{\textbf{\NoStateModels.}}
This design is built upon two conditions:
\begin{itemize}
    \item \textit{Relative end-effector (EEF) action space}~\cite{chi2024universal}: The visuomotor policies predict relative displacements of the end-effector based on the current observation. Among different action spaces, the relative EEF action space most naturally supports the generalization of policies.

    \item \textit{Full task observation:} Another key condition for effective \NoStateModels is to ensure sufficient task-relevant visual information, which we term \say{full task observation}. This enables visuomotor policies to fully \say{see} the task-relevant objects in the task.
\end{itemize}
As shown in Figure~\ref{fig:teaser}, under the relative EEF action space, we ensure full task observation with dual wide-angle \eefcams (field of view approximately $120^\circ\times120^\circ$) mounted on the top and bottom of the end-effector, which provides sufficient task-relevant visual information for \NoStateModels even in some complex scenarios.

This mechanism of \NoStateModels forces the policy to develop a deeper understanding of the task environment rather than simply memorizing the trajectories, thereby enabling \NoStateModels to achieve advantages that \YesStateModels cannot provide:
\begin{itemize}
    \item \textit{Spatial Generalization:} Since \NoStateModels do not rely on state inputs, they avoid overfitting to the training trajectories. 
    Therefore, they exhibit strong height and horizontal generalization abilities, where height refers to variations of the task-relevant object's location in the vertical direction, and horizontal refers to variations of the object's location in the 2D plane.

    \item \textit{Data efficiency:} Even in in-domain settings, \YesStateModels require diverse demonstrations to avoid overfitting to specific trajectories. In contrast, removing the state input eliminates this dependence on trajectory diversity, allowing \NoStateModels to be fine-tuned with less demonstration data. This reduces the cost of data collection, which is often a major bottleneck in deploying real-world robots.
\end{itemize}

% \newpage

\begin{itemize}
    \item \textit{Cross-embodiment adaptation:} Since \NoStateModels rely only on visual inputs and predict actions in the relative EEF space, they exhibit stronger cross-embodiment adaptation ability than \YesStateModels. They do not require additional adaptation to different state spaces, so the same task can be easily adapted to new embodiments with fewer fine-tuning steps.
\end{itemize}

We have conducted extensive experiments across a diverse range of tasks, robot embodiments, and policy architectures.
In both real-world and simulation environments, \NoStateModels achieve comparably great in-domain performance to \YesStateModels.
Most importantly, when trained on strictly collected real-world demonstration data
(i.e., the task-relevant object location has a constrained initial distribution range),
\NoStateModels exhibit significantly stronger spatial generalization ability than \YesStateModels.
For further benefits, e.g., data efficiency and cross-embodiment adaptation ability, they also demonstrate obvious advantages over \YesStateModels, highlighting their potential for scalable and practical deployment in real-world robotic systems.

% \NoStateModels exhibit significantly improved spatial generalization, data efficiency, and cross-embodiment adaptation ability than \YesStateModel.

%% file: Figures/1_teaser/teaser.tex
\begin{figure}[t]
    \centering
    \includegraphics[width=\columnwidth]{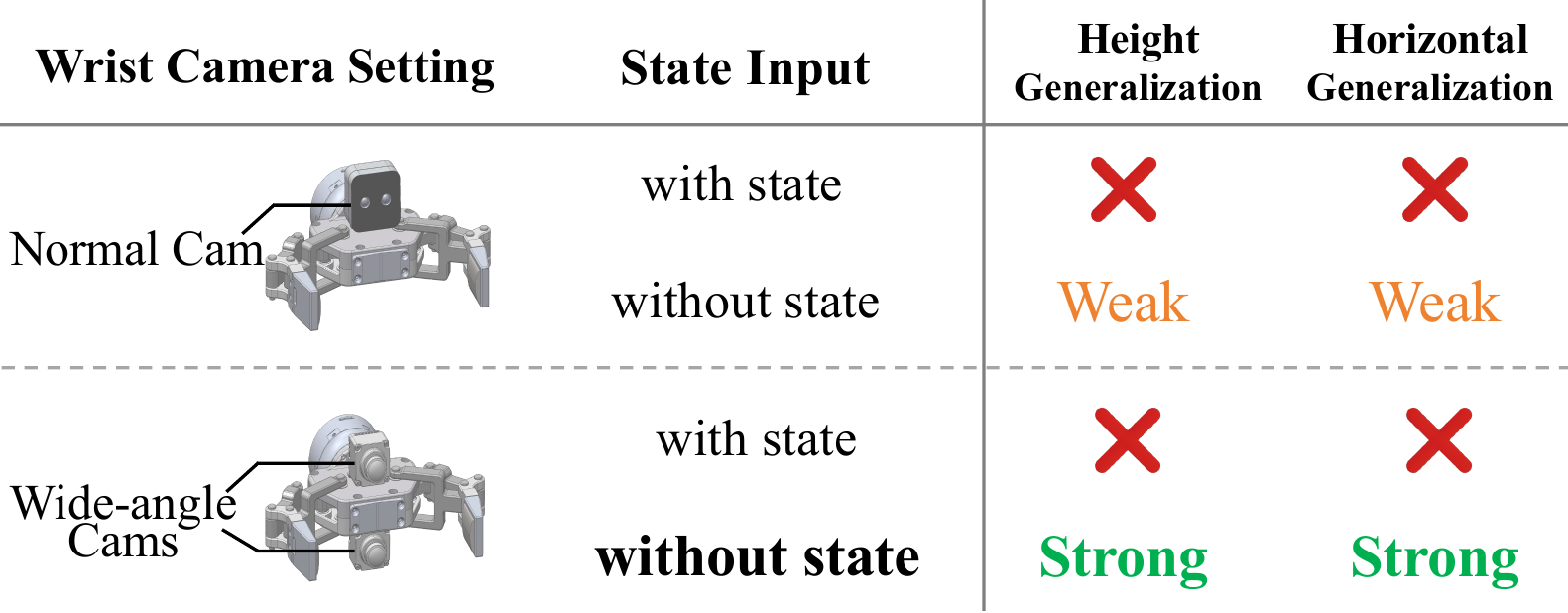}
    \caption{With relative EEF action space and full task observation, \NoStateModels demonstrate significantly improved spatial generalization ability compared to \YesStateModels. To handle complex scenarios, the full task observation is implemented by dual wide-angle \eefcams.}
    \label{fig:teaser}
\end{figure}

%% file: Sections/2_related_works.tex
\section{Related Works}
\textbf{Visuomotor Policies and State Inputs.}
Imitation-learning-based visuomotor policies~\cite{chi2023diffusion, zhao2023learning, black2410pi0, bjorck2025gr00t, liu2024rdt, zitkovich2023rt, brohan2022rt,team2024octo,kim2024openvla,barreiros2025careful,Liu2024RDT1BAD,Lin2025OneTwoVLAAU,Zhang2025KineDexLT,Wang2025VQVLAIV} have been widely adopted for robotic manipulation, achieving remarkable performance across diverse tasks. Recent advances such as ACT~\cite{zhao2023learning}, Diffusion Policy~\cite{chi2023diffusion}, and \pizero~\cite{black2410pi0} highlight the effectiveness of combining large-scale trajectory datasets with powerful model architectures.
A common practice in these approaches is to incorporate proprioceptive state inputs alongside visual observations to stabilize training and improve fine-tuning efficiency~\cite{torabi2019imitation, zhao2023learning, chi2023diffusion, black2410pi0, kuang2025adapt,kim2024openvla,barreiros2025careful}.
While such state inputs provide compact information about the robot’s configuration, they also create a shortcut for the policy: instead of reasoning from visual cues, the policy can simply memorize training trajectories tied to specific states~\cite{yang2025s, kuang2025adapt, chen2025improving, zeng2021transporter, yin2023spatial}.
As a result, the policy overfits the training data and cannot adapt to spatial layout changes, limiting its spatial generalization.

\textbf{Improving Spatial Generalization.}
Several efforts focus on improving the spatial generalization of visuomotor policies.
Data-driven approaches~\cite{Zhao2024ALOHAUA,o2024open} collect the demonstration data across diverse spatial configurations, but quickly become prohibitively expensive in the real world~\cite{mitrano2022data}.
Simulation methods~\cite{Zhao2020SimtoRealTI,Tobin2017DomainRF,Makoviychuk2021IsaacGH,OpenAI2019SolvingRC} still often struggle with the sim-to-real gap.
Representation-driven approaches build object-centric features to isolate task-relevant information, but often depend on complex perception pipelines~\cite{chapin2025object, emukpere2025disentangled, chen2025improving,hansen2021generalization,yen2020learning,mandi2022towards}.
Architectural approaches introduce physical symmetries to promote invariance, but such assumptions often break under real-world conditions~\cite{wang2022robot, seo2025se}.
Regularization approaches penalize absolute position encoding to encourage relational reasoning, but add training complexity with limited gains~\cite{yin2023spatial}.
While such methods can be effective, they introduce additional cost or require non-trivial algorithmic design, highlighting the need for a simpler and more practical solution to spatial generalization.

%% file: Sections/3_no_state.tex
\section{Methodology}
\subsection{Preliminary}
\subsubsection{\textbf{Imitation-Learning-Based Visuomotor Policies}}
We consider visuomotor policies mapping raw observations to low-level control actions.
At time $t$, the observation is $o_t \in \mathcal{O}$ (camera images and, conventionally, states), and the policy with trainable parameters $\theta$ is defined as: $\pi_\theta({a}_t \mid o_t)$, where $a_t$ denotes the low-level control action.
In imitation learning~\cite{bain1995framework}, the policy is trained on demonstration data $\mathcal{D}$ by minimizing the negative log-likelihood of actions:
\begin{equation}
    \mathcal{L}_{\text{IL}}(\theta) = -\sum_{(o_t, {a}_t) \in \mathcal{D}} \log \pi_\theta({a}_t \mid o_t).
\end{equation}
During deployment, $\pi_\theta$ takes online observations $o_t$ and outputs the actions ${a}_t$, which are executed on the robot.

\subsubsection{\textbf{Action Representation Space}}
We consider two common action representation spaces: relative EEF action and relative joint-angle action.

In the relative EEF action space, the end-effector pose at time $t$ is
${p}_t = \big[ {x}_t, {q}_t \big]$, where ${x}_t \in \mathbb{R}^3$ is the Cartesian position and ${q}_t \in SO(3)$ is the orientation.
The policy outputs a relative displacement:
\begin{equation}
    \label{eq:delta_eef}
    {a}_t = \Delta {p}_t = \big[ \Delta {x}_t, \Delta {q}_t \big],
\end{equation}
where $\Delta {x}_t$ and $\Delta {q}_t$ denote the translation and rotation.
The next end-effector pose is updated by: ${p}_{t+1} = {p}_t \oplus \Delta {p}_t$, where $\oplus$ denotes composition of the translation and rotation.

In the relative joint-angle action space, the policy drives the end-effector motion by predicting relative joint changes $\Delta \theta_t$.
In this case, the end-effector displacement depends on both $\Delta \theta_t$ and the current joint pose $\theta_t$, i.e.,
\begin{equation}
    \label{eq:delta_joint}
    \Delta {p}_t = f(\Delta \theta_t, \theta_t),
\end{equation}
where $f$ denotes the forward kinematics mapping.

\input{Tables/state_challenge_section3}

\subsection{Spatial Generalization Challenge of State Input}
Proprioceptive states provide direct and accurate robot configuration, but may act as shortcuts, where the policy directly associates absolute states with expert trajectories.
Consequently, the policy tends to overfit to the training trajectories and fails to adapt to spatial layout changes.

We validate this limitation with a real-world height generalization evaluation as an example.
Specifically, we collect \say{\textit{Pick a Pen into Pen Holder}} demonstrations (the overview is illustrated in Figure~\ref{fig:demo}) at a fixed 80 cm table height and fine-tune the \pizero~\cite{black2410pi0} policy using the relative EEF action space.
As shown in the first 3 columns of Table~\ref{tab:state_challenge_sec3}, \YesStateModels completely fail to generalize across different table heights.
However, adding simple hacks (manually shifting the state according to table height) effectively improves the height generalization ability, indicating that state input is a critical factor limiting spatial generalization.
At the same time, adding random noise augmentation ($[-5\ \text{cm},\ 5\ \text{cm}]$) on the state height dimension $z$ improves height generalization without affecting in-domain performance, indicating that state input may not be necessary. These motivate us to consider removing the state input altogether.

\subsection{\NoStateModels}
In this study, we propose to remove the state input from visuomotor policies, based on the \textbf{relative EEF action space} and \textbf{full task observation}.
This improves the spatial generalization without requiring additional architectural changes or costly diverse data collection.

\subsubsection{\textbf{Relative EEF Action Space}}
\label{sec:discuss_action_space}
To study spatial generalization, we begin by clarifying the action space.
First, we exclude all absolute actions, where the policy predicts absolute poses, simply learning fixed mappings tied to training trajectories and thus failing to adapt to new spatial layouts.
Here, we consider two common relative action spaces: relative EEF action and relative joint-angle action:

\begin{itemize}
    \item \textit{Relative Joint-angle Action Space:}
    In spatial generalization tasks such as height generalization, as the table height changes, when the end-effector is at the same relative position with respect to the table, the robot receives the same visual observations. In this case, the policy predicts the same $\Delta q_t$, but the joint configurations ${q}_t$ are different. As shown in Eq.~\ref{eq:delta_joint}, this results in different end-effector displacements $\Delta p_t$, leading to incorrect actions.

    \item \textit{Relative EEF Action Space:}
    As shown in Eq.~\ref{eq:delta_eef}, the policy predicts relative end-effector motions directly from observations. The action $\Delta p_t$ depends only on the observations, not on the absolute pose, so identical observations yield the same displacement regardless of absolute robot poses. This invariance allows relative EEF actions to support spatial generalization across heights and horizontal positions.
\end{itemize}
Since the relative EEF action space naturally supports the spatial generalization, our \NoStateModels will be built upon this action representation space.

\subsubsection{\textbf{Full Task Observation}}
\label{sec:discuss_full_observation}
A key condition for \NoStateModels is \say{full task observation}, ensuring the policy receives sufficient visual information.
With state input, the policy can directly learn shortcut associations, such as what action to take once the robot reaches a certain configuration, without relying on sufficient visual information.
In contrast, without state input, the policy has to make decisions entirely from visual information, which requires providing full task-relevant visual observations, i.e., the full task observation.
This motivates us to equip the end-effector with a broader field of view for a wide range of scenarios.

\input{Figures/2_cam/cam}

Our camera system consists of an overhead camera and \eefcams.
In the normal \eefcam setting, a single normal-view \eefcam is mounted on top of the end-effector (in this study with view field $87^\circ\times58^\circ$), as illustrated in Figure~\ref{fig:single_and_dual_cam}\textcolor{red}{(a)}.
To achieve full task observation, we adopt dual wide-angle \eefcams (field of view $120^\circ\times120^\circ$) mounted on the top and bottom of end-effectors, as illustrated in Figure~\ref{fig:single_and_dual_cam}\textcolor{red}{(b)}.
This setting expands the view and exposes the workspace beneath the end-effector (note that in tasks with simple scenarios, e.g., involving a single task-relevant object, the normal \eefcam setting can already be enough).

\input{Figures/demo/demo}

\subsubsection*{\textbf{Summary}}  
As shown in the last 2 columns of Table~\ref{tab:state_challenge_sec3}, in simple attempts at height generalization on task \textit{\say{Pick Pen}}, the relative EEF action space and full task observation provide the improved height generalization of \NoStateModels, significantly outperforming \YesStateModels.
In the following sections, we will conduct extensive evaluations to validate and further analyze \NoStateModels.
Meanwhile, we demonstrate their further benefits, including higher data efficiency and better cross-embodiment adaptation.
In addition, we also demonstrate an interesting finding that removing the overhead camera can further enhance the policy's spatial generalization ability.

%% file: Tables/state_challenge_section3.tex
\begin{table*}[t]
    \centering
    \caption{Success rates in the exploration of table height generalization, using the ``\textit{Pick a Pen into Pen Holder}'' task as the representative example. Simple hacks and noise on the state improve the spatial generalization, indicating the state as a bottleneck. This motivates removing the state input, as confirmed by empirical results.}
    \begin{tabular}{c c c c :c c}
        \toprule
        Wrist-camera setting        &   Normal wrist-cam   &   Normal wrist-cam   &   Normal wrist-cam   &   Normal wrist-cam   &   Dual wide-angle wrist-cams \\
        Method on state             &     w/ state     &     State hack     & Noise augmentation &    w/o state   &        w/o state       \\
        \midrule
        $h$=80cm (In-domain)     &      30/30         &      30/30         &       30/30        &       30/30        &   \textbf{30/30}   \\
        $h$=72cm (Out-of-domain) &       0/30         &      30/30         &       24/30        &       30/30        &   \textbf{30/30}   \\
        $h$=90cm (Out-of-domain) &       0/30         &      30/30         &       11/30        &       17/30        &   \textbf{29/30}   \\
        \bottomrule
    \end{tabular}
    \label{tab:state_challenge_sec3}
\end{table*}

%% file: Figures/2_cam/cam.tex
% \begin{figure}[htbp]
%   \centering
%   \begin{minipage}{0.48\linewidth}
%     \centering
%     \includegraphics[width=\linewidth]{Figures/2_cam/single.png}
%     \caption{Single normal camera setting. Sometimes the target may not be visible, such as the pen holder.}
%     \label{fig:single_cam}
%   \end{minipage}\hfill
%   \begin{minipage}{0.48\linewidth}
%     \centering
%     \includegraphics[width=\linewidth]{Figures/2_cam/dual.png}
%     \caption{Dual wide-angle cameras setting. It provides complete and sufficient observations of the task.}
%     \label{fig:dual_fisheye}
%   \end{minipage}
% \end{figure}

% [Review 版]
% \begin{figure}[htbp]
\begin{figure}[t]
    \centering
    \includegraphics[width=\columnwidth]{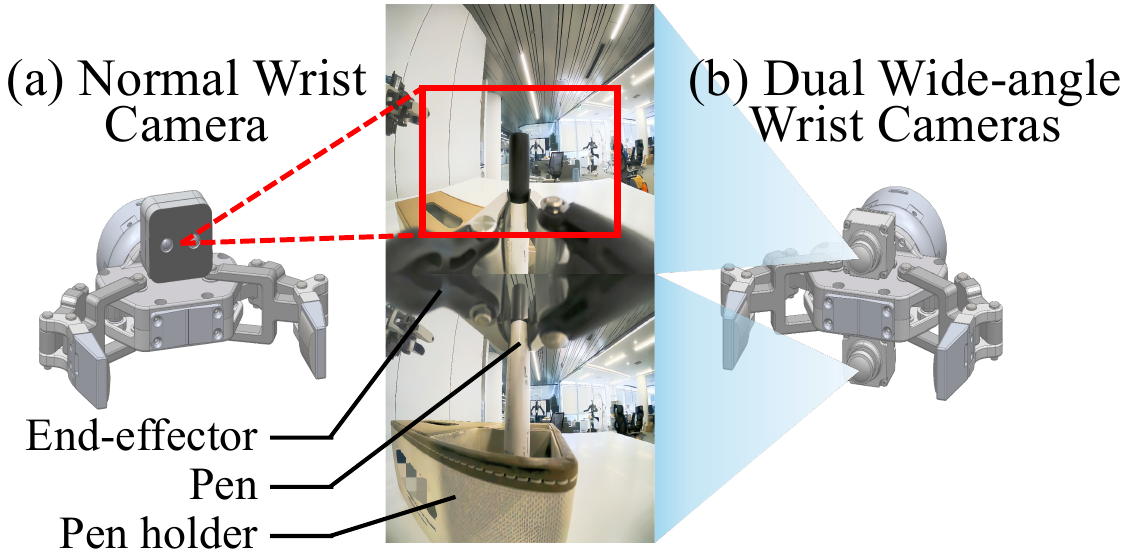}
    \caption{(a) Normal \eefcam setting. Sometimes the target may not be visible. (b) Dual wide-angle \eefcams setting. It provides sufficient observations of the task.}
    \label{fig:single_and_dual_cam}
\end{figure}

%% file: Figures/demo/demo.tex
\begin{figure*}[t]
    \centering
    \includegraphics[width=\textwidth]{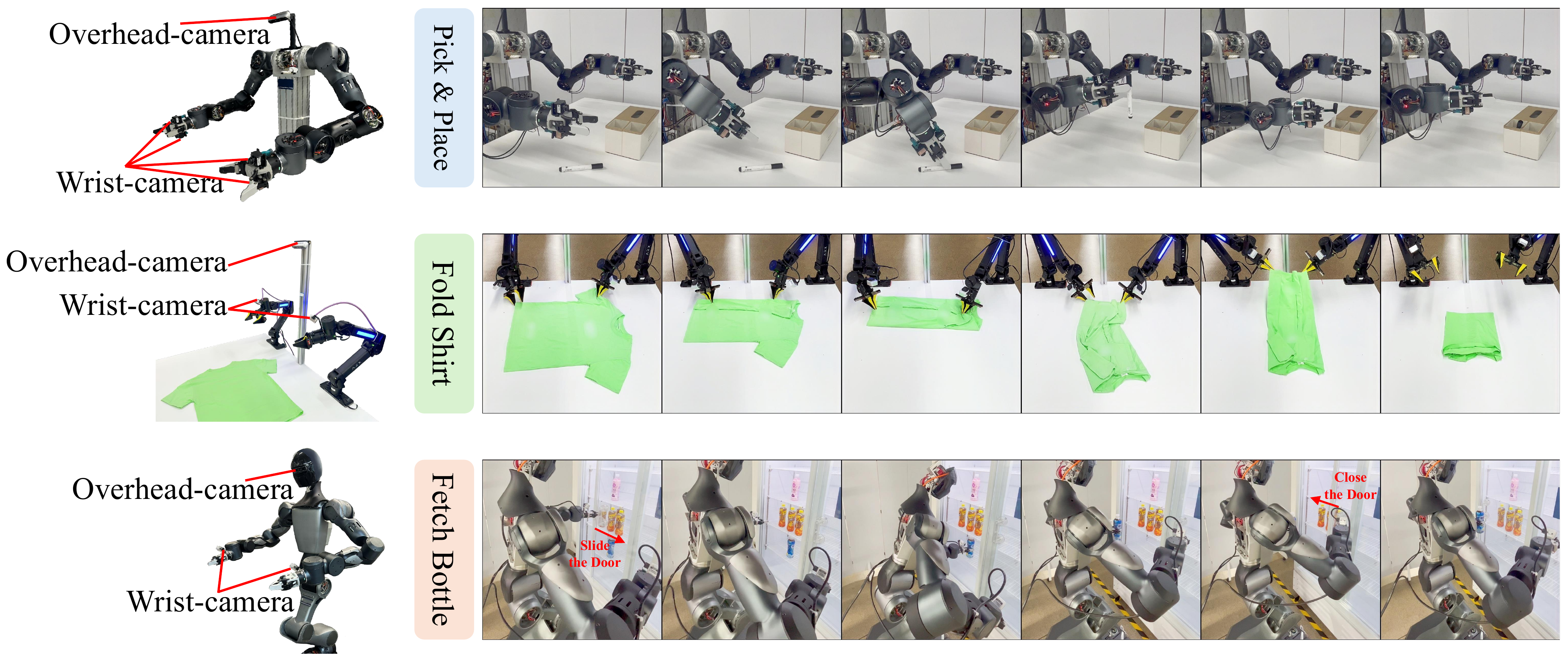}
    \caption{Overview of our robot embodiments and representative tasks, including ``\textit{Pick $\&$ Place}'' tasks, more challenging ``\textit{Fold Shirt}'' and ``\textit{Fetch Bottle}'' task. These tasks span a wide range of robot embodiments: a $2\times8$ DoF human-like dual-arm robot, a $2\times7$ DoF dual-arm Arx5 robotic arm system, and a 26 DoF whole-body robot.}
    \label{fig:demo}
\end{figure*}

%% file: Sections/4_experiments.tex
\section{Performance Across Various Tasks}
To evaluate the performance of \NoStateModels, we conduct extensive evaluations across various tasks.
Performances of the \YesStateModel and several optimizing strategies on it can be found in Appendix Section~\ref{sec:app_state_challenge}.

\subsection{Setup}
\subsubsection{\textbf{Task}}
Our real-world tasks include 3 \textit{\pickplace} tasks, a more challenging \textit{\say{Fold Shirt}} task, and a complex task \textit{\say{Fetch Bottle}} (on a whole-body robot with torso, waist, and leg motions controlled by a 6-dimensional torso pose vector, representing position and orientation in the EEF form).
The detailed task descriptions are as follows:

\begin{itemize}
    \item \textit{Pick Pen}: Pick up a pen from the table and place it into a pen holder on the table.
    \item \textit{Pick Bottle}: Grasp the bottle cap and remove the bottle from the step on the table.
    \item \textit{Put Lid}: Pick up the lid from the table and accurately place it on a large teacup on the table.
    \item \textit{Fold Shirt}: Fold the shirt that is laid flat on the table.
    \item \textit{Fetch Bottle (whole-body)}: Open the refrigerator door, take out the bottle, and close the refrigerator door.
\end{itemize}
As shown in Figure~\ref{fig:demo}, we present the overviews of the robot embodiments and representative tasks in our evaluations.

We also conduct evaluations in the simulation environment on the LIBERO benchmark~\cite{Liu2023LIBEROBK}, where we fine-tune the policy separately in each suite and subsequently evaluate it in the corresponding test suite.

\subsubsection{\textbf{Real-world Data}}
We employ professional data collectors to collect the real-world demonstration data using the teleoperation.
For each \textit{\pickplace} task, 300 trajectory episodes, around 5 hours of data, are collected.
For the challenging \textit{\say{Fold Task}} and \textit{\say{Fetch Bottle (whole-body)}} task, 10,000 episodes, around 80 hours of data, are collected.

Importantly, during data collection, we fix the table height and limit the task-relevant object locations within a constrained 2D range.
Taking the task \textit{\say{Pick Pen}} as an example, as shown in Figure~\ref{fig:range}, in training data, the pen holder location is fixed, and in horizontal evaluation, we shift its location.
This design ensures that the spatial generalization ability originates from the policy itself rather than from diverse data.
% (Note that even under such narrow location variations, multiple trajectories are still required to prevent overfitting to specific trajectories, especially for policies with state input.)

\input{Figures/range/range}

\input{Figures/Base_Experiments/pick_place}

\subsubsection{\textbf{Evaluation Metric}}
\label{sec:exp_metric}
We evaluate the spatial generalization of visuomotor policies along two dimensions: height and horizontal generalization.
Each real-world evaluation consists of 30 trials, with success counted only if the entire trajectory is completed.
A trial is marked as a failure if the policy takes no reasonable action within 30 seconds or if any action fails.

\paragraph{Height Generalization Evaluation}
The \textit{\pickplace} data are collected at 80 cm table height.
The height generalization score is computed as the average success rate of the total 60 trials at 72 cm and 90 cm table heights.
And as shown in Figure~\ref{fig:demo}, since the Arx5 arms are fixed to the table and the refrigerator height cannot be adjusted, the \textit{\say{Fold Shirt}} and \textit{\say{Fetch Bottle (whole-body)}} tasks are not applicable for height generalization evaluation.

\paragraph{Horizontal Generalization Evaluation}
In the \textit{\pickplace} and \textit{\say{Fetch Bottle (whole-body)}} tasks, the target objects (pen holder, step, large teacup, and refrigerator) are shifted within 2D ranges of 5 cm and 10 cm, as illustrated by the \textit{\say{Pick Pen}} example in Figure~\ref{fig:range}.
In the \textit{\say{Fold Shirt}} task, we evaluate by laterally shifting a single Arx5 arm by 15 cm, as well as shifting both arms simultaneously by 15 cm in opposite directions.
Each task is evaluated with 60 trials (two runs of 30), and the horizontal generalization score is the average success rate across them.

\subsubsection{\textbf{Model}}
In our main evaluations, we use \pizero~\cite{black2410pi0} for imitation learning, following its released fine-tuning recipe.
\pizero is widely regarded as one of the most powerful visuomotor policies in the community.
In addition, our detailed analysis in Section~\ref{sec:exp_model_structure} further evaluates different policy architectures, including ACT (Action-conditioned Transformer)~\cite{zhao2023learning}, which employs action-conditioned attention to capture temporal dependencies, and Diffusion Policy~\cite{chi2023diffusion}, which models actions as a distribution via diffusion dynamics.

% Note that due to hardware limitations, both the Arx5 arm and the full-body robot are unable to mount the dual wide-angle cameras on the end-effectors.
% However, since the scenarios are relatively simple, experimental results show that using the normal camera settings still provides full task observation, ensuring that the \NoStateModels can successfully complete the tasks.

\subsection{Real-world Evaluations}
\label{sec:base_exp}
Here we report the spatial generalization evaluations, including height and horizontal generalization, on 5 real-world tasks.
And we will report their in-domain performance and the simulation evaluations in Appendix Section~\ref{sec:app_indomain_exp}.

In Figure~\ref{fig:base_exp_pick_and_place}, we report the height and horizontal generalization performance in 3 real-world \textit{\pickplace} tasks.
Compared to \YesStateModels, \NoStateModels exhibit significant improvement in both height and horizontal generalization:
taking the \textit{\say{Pick Pen}} task as an example, the success rate in height generalization rises from 0 to 0.98, and in horizontal generalization from 0 to 0.58.
And compared with the normal \eefcam setting, the height generalization success rate improves from 0.87 to 0.98, and the horizontal generalization from 0.27 to 0.58.

\input{Tables/exp_shirt_and_fullbody}

In task \textit{\say{Fold Shirt}}, folding a shirt is difficult due to the deformable nature of fabric which makes the folding manipulation challenging.
And task \textit{\say{Fetch Bottle (whole-body)}} is more difficult because the robot's torso motions are not directly observable.
As discussed in Section~\ref{sec:exp_metric}, height generalization evaluation is not applicable to these two tasks.
In addition, due to hardware limitations, the dual wide-angle \eefcams cannot be mounted in these embodiments.
In Table~\ref{tab:exp_shirt_and_fullbody}, we report the horizontal generalization performance on these two tasks.
Even in these complex and challenging tasks, \NoStateModels still achieve significantly stronger spatial generalization ability.
Moreover, this also reflects that for simpler scenarios (i.e., with simple task-relevant objects), the normal \eefcam setting can still provide full task observation.

\section{Detailed Analysis on \NoStateModels}
For deeper insights, we conduct a more detailed analysis of \NoStateModels, examining how their behavior varies under different conditions.
In this section, we mainly focus on the \textit{\say{Pick Pen}} task as the example, which is both intuitive and typical, making it well-suited for detailed analysis.
Unless otherwise specified, all evaluations in this section use the \pizero policy and the relative EEF action space.

% This fine-grained investigation allows us to better understand the factors that contribute to their robustness and generalization.

% In this section, we mainly focus on the \textit{\say{Pick Pen}} task as the example, which is both intuitive and typical, making it well-suited for detailed analysis.
% While when evaluating the ability of cross-embodiment adaptation, we use the \textit{\say{Fold Shirt}} task.
% We continue to employ the \pizero policy with relative EEF action space, and we will conduct evaluations across different action spaces and policy architectures in Section~\ref{sec:exp_action_representation} and~\ref{sec:exp_model_structure}.

\subsection{Action Representation}
\label{sec:exp_action_representation}

\input{Tables/action_space}
As discussed in Section~\ref{sec:discuss_action_space}, the relative EEF action space most naturally supports the generalization ability of \NoStateModels.
In this section, we evaluate alternative action representations, including the absolute EEF, both absolute and relative joint-angle action spaces.
Evaluations are performed using the dual wide-angle \eefcams setting.

As reported in Table~\ref{tab:action_space}, the relative EEF action space achieves the best performance in both in-domain and spatial generalization settings, whereas others show disastrous performance in spatial generalization.
These results highlight that the relative EEF action space most naturally supports the generalization ability of \NoStateModels.

\subsection{Full Task Observation}

\input{Tables/full_observation}

% \NoStateModels require full task observation for stronger spatial generalization.
Another key condition for stronger spatial generalization in \NoStateModels is full task observation.
Our camera system includes an overhead camera and \eefcams.
As illustrated in Figure~\ref{fig:single_and_dual_cam}, each end-effector is equipped with two wide-angle \eefcams on the top and bottom.
By cropping image regions or masking one of the inputs, we create different levels of task observation to demonstrate the critical role of full task observation in \NoStateModels.

As reported in Table~\ref{tab:full_observation}, the spatial generalization ability of \NoStateModels gradually improves as the field of view expands, indicating that full task observation is important for achieving strong generalization in \NoStateModels.

Moreover, an interesting finding shows that even without the overhead camera, the dual wide-angle \eefcams alone enable the best spatial generalization. This indicates that, in the current task, they provide completely full task observation for the entire trajectory, while the overhead camera is not only unnecessary but can even be harmful (we will discuss further in Section~\ref{sec:exp_new_sensor}).

\subsection{Policy Architecture}
\label{sec:exp_model_structure}

\input{Tables/model_structure}
In addition to \pizero policy, we also evaluate other policiy architectures without state input, including ACT and Diffusion Policy.
All use the dual wide-angle \eefcams setting.

As reported in Table~\ref{tab:model_structures}, the results are consistent across architectures: \NoStateModels exhibit much stronger spatial generalization than \YesStateModels, indicating their effectiveness is independent of specific policy implementations, representing a general and universal conclusion.

\section{Further Benefits of \NoStateModels}
In this section, we will demonstrate additional potential advantages of \NoStateModels.

% \subsection{Higher Data Efficiency}
% \label{sec:exp_less_data}

% \input{Tables/data_scale}

% To avoid overfitting, policies with state input require more demonstration data for fine-tuning, even in-domain.
% In contrast, \NoStateModels depend less on trajectory diversity and can be fine-tuned with fewer data.
% We evaluate on the in-domain \textit{\say{Pick Pen}} task using dual wide-angle end-effector cameras, varying fine-tuning data to 200, 100, and 50 episodes, and measuring performance after fine-tuning 2 and 4 epochs.
% As shown in Table~\ref{tab:data_scale}, reducing fine-tuning data causes state-based policies to overfit to specific trajectories and lose success, but \NoStateModels maintain much higher performance.

\input{Figures/Base_Experiments/data_scale}

\subsection{Higher Data Efficiency}
\label{sec:exp_less_data}

% Even in in-domain settings, policies with state input require diverse demonstrations to avoid overfitting to specific trajectories, greatly raising data collection costs. \NoStateModels mitigate this issue: without state input, policies are less prone to memorizing trajectories and can achieve comparable performance with fewer fine-tuning data, thereby enhancing data efficiency and practicality for real-world deployment.

Even in in-domain settings, \YesStateModels require diverse demonstrations to avoid overfitting to specific trajectories, greatly raising data collection costs. While, \NoStateModels are less prone to memorizing specific trajectories and can achieve comparable performance with fewer fine-tuning data, thereby enhancing data efficiency and practicality for their real-world deployment.

We validate this on the in-domain \textit{\say{Pick Pen}} task with dual wide-angle \eefcams, varying fine-tuning data to 300, 200, 100, and 50 episodes, and measuring after 2 and 4 fine-tuning epochs. As shown in Figure~\ref{fig:data_scale}, reducing data leads state-based policies to overfit and lose success, while \NoStateModels maintain much higher performance.

\subsection{Better Cross-embodiment Adaptation}
\label{sec:exp_cross_embodiment}
We find that \NoStateModels also benefit the cross-embodiment fine-tuning. For \YesStateModels, cross-embodiment adaptation requires aligning with a new state space, and even with EEF-based states, differences in reference frame definitions across embodiments still create gaps. In contrast, \NoStateModels avoid this issue: with similar camera setups, they only adapt to minor image shifts, enabling more efficient cross-embodiment fine-tuning.

\input{Tables/cross_embodiment}

We validate this on the \textit{\say{Fold Shirt}} task (in-domain setting). Policies are first trained on dual-arm Arx5 (the EEF space is in table frame) and then adapted to a human-like dual-arm robot (the EEF space is in robot-centric frame). We collect 100 shirt-folding demonstrations on the human-like robot and fine-tune the \pizero policy with and without state input, each initialized from its corresponding Arx5 checkpoint. As shown in Table~\ref{tab:cross_embodiment}, \NoStateModels adapt much faster across embodiments, achieving substantially higher success rates than state-based policies under the same fine-tuning epochs.
This indicates that \NoStateModels have a better cross-embodiment ability than \YesStateModels.

\section{Rethinking the Overhead Camera}
\label{sec:exp_new_sensor}
After removing the state input that limits spatial generalization, we consider the overhead camera might as another potential bottleneck.
Changes in object locations can induce distribution shifts in overhead camera images, and in extreme cases (e.g., 100 cm table height) degrade performance.
In contrast, since the end-effector can move along with the task-relevant object, the \eefcam can still capture observations consistent with those in training, avoiding the out-of-domain issues.
Given that the dual wide-angle \eefcams already provide full task observation, the overhead camera may not only be unnecessary but even harmful.

\input{Tables/no_high}

We evaluate this through experiments on the \textit{\say{Pick Pen}} task under more challenging scenarios:
\begin{itemize}
    \item Raising the table height to 100 cm.
    \item Raising the pen holder to double its height, changing its relative hieght with respect to the table.
    \item Moving the pen holder 20 cm away from its position in training data.
\end{itemize}

As reported in Table~\ref{tab:no_high}, \NoStateModels with the overhead camera show terrible performance across all 3 more challenging scenarios.
While without the overhead camera, success rates remain consistently high, confirming that dual wide-angle \eefcams alone are sufficient, while the overhead view introduces harmful shifts.
This finding motivates us to rethink sensor design, perhaps removing the overhead camera, for future visuomotor policies.

%% file: Figures/range/range.tex
\begin{figure}[t]
    \centering
    \includegraphics[width=\columnwidth]{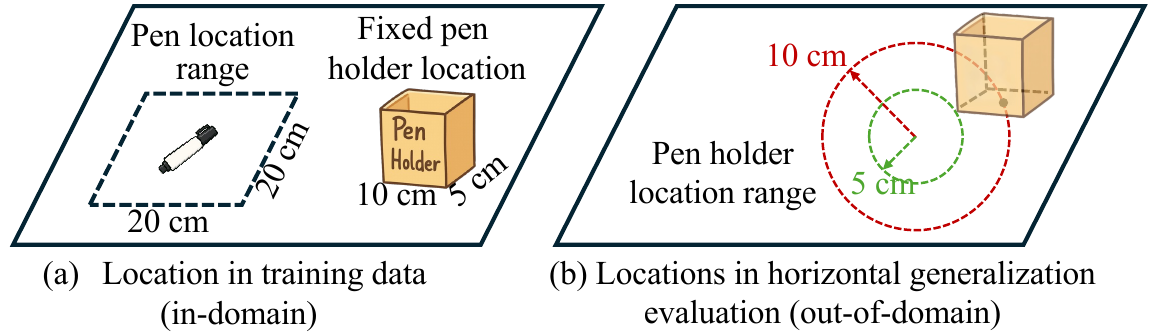}
    \caption{Task-relevant object locations illustration in task \textit{``Pick Pen''}. During training, the pen holder is fixed. For horizontal generalization, we keep the pen location range unchanged and shift the pen holder by 5 cm and 10 cm to compute the average success rate.}
    \label{fig:range}
\end{figure}

%% file: Figures/Base_Experiments/pick_place.tex
% \begin{figure*}[t]
%   \centering
%   % ===== Row 1 =====
%   \begin{subfigure}[t]{0.32\textwidth}
%     \centering
%     \includegraphics[width=\linewidth]{Figures/Base_Experiments/pick_bottle.pdf}
%     \caption{Pick Bottle}
%     \label{fig:exp_pick_bottle}
%   \end{subfigure}\hfill
%   \begin{subfigure}[t]{0.32\textwidth}
%     \centering
%     \includegraphics[width=\linewidth]{Figures/Base_Experiments/pick_pen.pdf}
%     \caption{Pick Pen}
%     \label{fig:exp_pick_pen}
%   \end{subfigure}\hfill
%   \begin{subfigure}[t]{0.32\textwidth}
%     \centering
%     \includegraphics[width=\linewidth]{Figures/Base_Experiments/put_lid.pdf}
%     \caption{Put Lid}
%     \label{fig:exp_put_lid}
%   \end{subfigure}
%   \caption{The height generalization and horizontal generalization (written as Height Gen. and Horizontal Gen.) performance across 3 real-world ``Pick $\&$ Place'' manipulation tasks.}
%   \label{fig:base_exp_pick_and_place}
% \end{figure*}

\begin{figure*}[t]
    \centering
    \includegraphics[width=\textwidth]{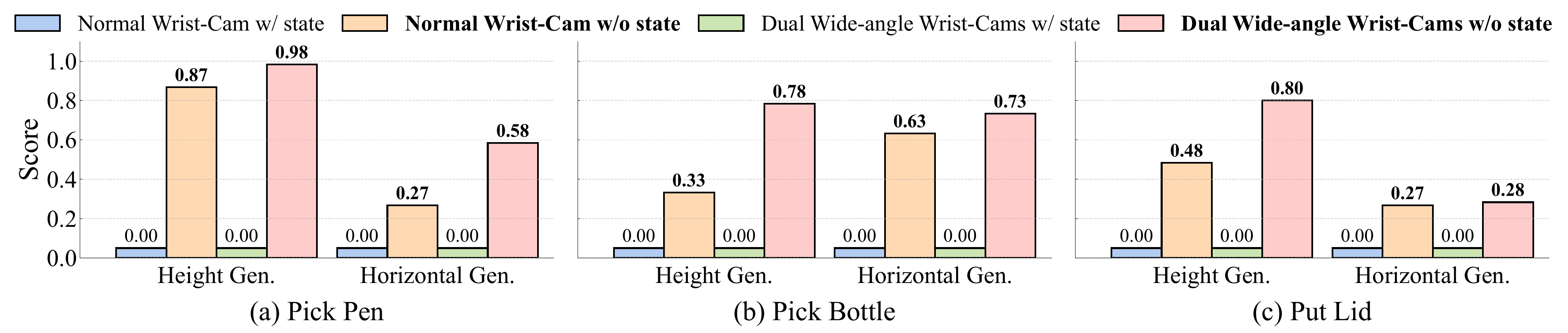}
    \caption{The height and horizontal generalization (written as Gen.) performances across 3 real-world \textit{``Pick $\&$ Place''} tasks. With full task observation, \NoStateModels show significantly improved spatial generalization than \YesStateModels.}
    \label{fig:base_exp_pick_and_place}
\end{figure*}

%% file: Tables/exp_shirt_and_fullbody.tex
\begin{table}[htbp]
  \centering
  \caption{The horizontal generalization performances across 2 challenging real-world tasks, \textit{``Fold Shirt''} and \textit{``Fetch Bottle (whole-body)''}. \NoStateModels still show significantly improved performance than \YesStateModels.}
  \label{tab:exp_shirt_and_fullbody}
  \begin{tabular}{c c c}
    \toprule
    Task name & \textit{Fold Shirt} & \textit{Fetch Bottle} \\
    \midrule
    w/ state (normal wrist-camera)  & 0.183          & 0.117          \\
    w/o state (normal wrist-camera) & \textbf{0.834} & \textbf{0.784} \\
    \bottomrule    
    % Task Name & \makecell{Normal Wrist-Camera \\ w/ state} & \makecell{Normal Wrist-Camera \\ w/o state} \\
    % \midrule
    % Fold Shirt               & 0.183 & \textbf{0.834} \\
    % Fetch Bottle             & 0.117 & \textbf{0.784} \\
    % \bottomrule
  \end{tabular}
\end{table}

%% file: Tables/action_space.tex
\begin{table}[htbp]
  \centering
  \caption{Spatial generalization evaluation of \NoStateModels using different action representations.}
  \label{tab:action_space}
  \begin{tabular}{lccc}
    \toprule
    Action space & In domain & Height Gen. & Horizontal Gen. \\
    \midrule
    Relative EEF            & 1.0 & 0.984 & 0.584 \\
    Absolute EEF         & 1.0 &  0   &  0   \\
    Relative joint-angle & 1.0 &  0   &  0   \\
    Absolute joint-angle & 1.0 &  0   &  0   \\
    \bottomrule
  \end{tabular}
\end{table}

%% file: Tables/full_observation.tex
\begin{table}[htbp]
    \centering
    \caption{Spatial generalization performances of \NoStateModels on different task observation levels, implemented by varying the camera settings.}
    \label{tab:full_observation}
    \begin{tabular}{@{}ccc:cc@{}}
    \toprule
        \begin{tabular}[c]{@{}c@{}}Wrist-cam\\ num\end{tabular} & \begin{tabular}[c]{@{}c@{}}Wrist-cam\\ type\end{tabular} & \begin{tabular}[c]{@{}c@{}}Overhead\\ cam\end{tabular} & Height Gen.  & Horizontal Gen. \\ \midrule
        N/A           & N/A                 & \checkmark   & 0.217        & 0.133           \\
        Single        & Normal              & \checkmark   & 0.867        & 0.267           \\
        Dual          & Normal              & \checkmark   & 0.920        & 0.400           \\
        Single        & Wide-angle          & \checkmark   & 0.917        & 0.500           \\
        Dual          & Wide-angle          & \checkmark   & 0.983        & 0.583           \\
        \textbf{Dual} & \textbf{Wide-angle} & \textbf{N/A} & \textbf{1.0} & \textbf{1.0}    \\ \bottomrule
    \end{tabular}
\end{table}

%% file: Tables/model_structure.tex
\begin{table}[htbp]
  \centering
  \caption{Spatial generalization comparison of policies with and without state input, using different model structures.}
  \label{tab:model_structures}
  \begin{tabular}{lcccc}
    \toprule
    Model structure & State input & Height Gen. & Horizontal Gen. \\
    \midrule
    \multirow{2}{*}{\pizero} & w/ state & 0 & 0 \\
                          & w/o state & 0.984 & 0.584 \\
    \midrule
    \multirow{2}{*}{ACT} & w/ state & 0 & 0.084 \\
                         & w/o state & 0.933 & 0.517 \\
    \midrule
    \multirow{2}{*}{Diffusion Policy} & w/ state & 0  &  0 \\
                                       & w/o state & 0.867  &  0.533 \\
    \bottomrule
  \end{tabular}
\end{table}

%% file: Figures/Base_Experiments/data_scale.tex
\begin{figure}[htbp]
    \centering
    \includegraphics[width=\columnwidth]{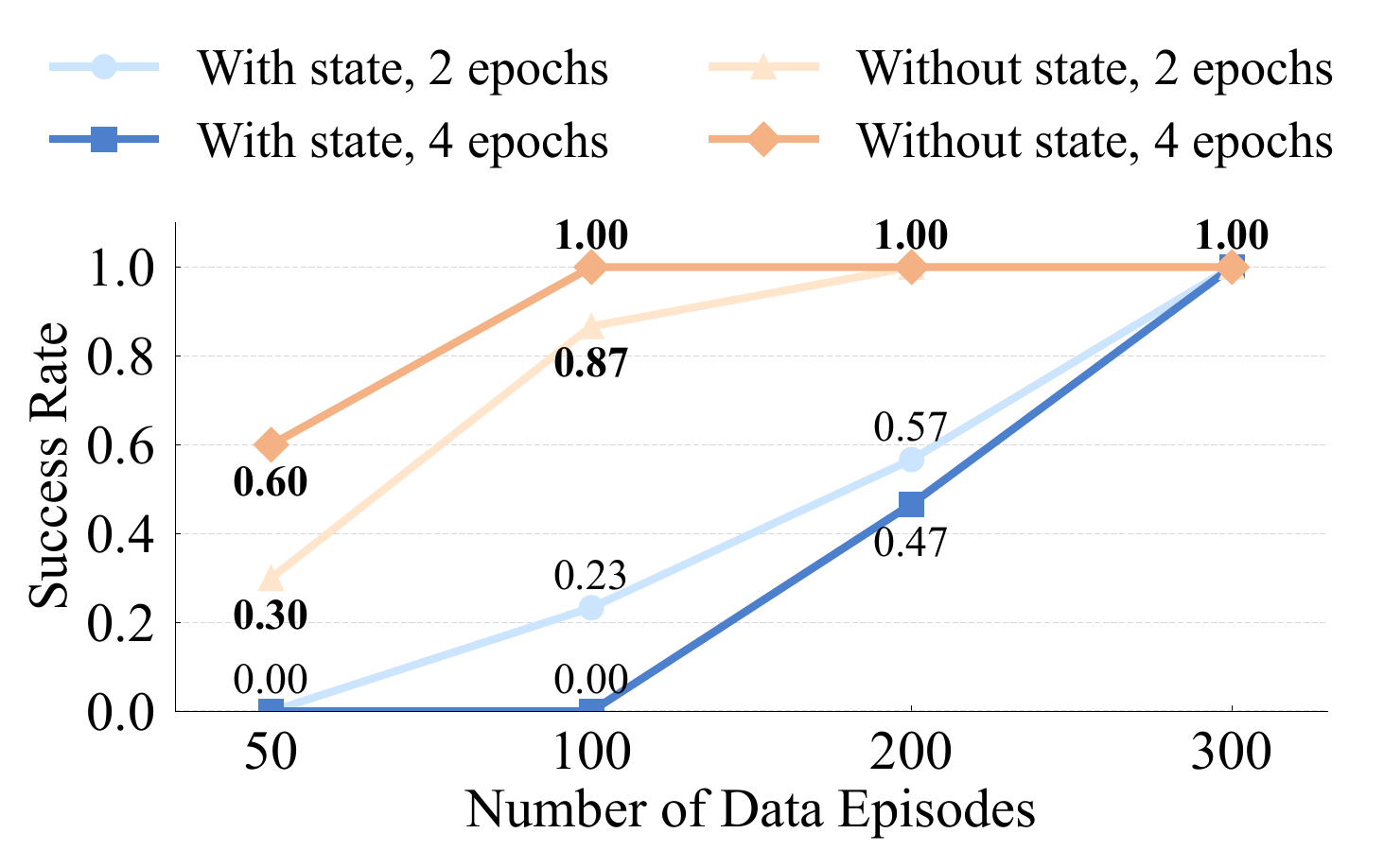}
    \caption{Evaluation success rates (in-domain) on the \textit{``Pick Pen''} task with varying amounts of fine-tuning data.}
    \label{fig:data_scale}
\end{figure}

%% file: Tables/cross_embodiment.tex
\begin{table}[htbp]
    \centering
    \caption{Success rates in in-domain \textit{``Fold Shirt''} task using the human-like robot. Each policy is fine-tuned from its corresponding checkpoint trained on Arx5 arms.}
    \begin{tabular}{c c c}
        \toprule
        State input & Fine-tune 5k steps & Fine-tune 10k steps \\
        \midrule
        w/ state    &  0.333  &  0.767  \\
        w/o state &  0.700  &  0.967  \\
        \bottomrule
    \end{tabular}
    \label{tab:cross_embodiment}
\end{table}

%% file: Tables/no_high.tex
\begin{table}[htbp]
    \centering
    \caption{Success rates with and without the overhead camera in more challenging \textit{``Pick Pen''} generalization scenarios, with dual wide-angle \eefcams.}
    \begin{tabular}{c c c c}
        \toprule
        Overhead camera & Table height & Raising pen & Moving pen        \\
        input      & 100cm        & holder height  & holder 20 cm \\
        \midrule
        w/ overhead cam    &  0  &  0.467  &  0  \\
        w/o overhead cam &  1.0 &  0.867  &  0.800 \\
        \bottomrule
    \end{tabular}
    \label{tab:no_high}
\end{table}

%% file: Sections/5_conclusion.tex
\section{Conclusion}
In this study, we propose the \NoStateModels, under two conditions: the relative EEF action space and the full task observation through sufficiently comprehensive visual information.
Without state input, these policies maintain perfect in-domain performance while achieving significant improvements in spatial generalization.
\NoStateModels also reduce the costly real-world data need, enable more efficient cross-embodiment adaptation, and inspire new directions in future sensor design.
Our findings shed new light on how \NoStateModels can serve as a foundation for building more generalizable robotic learning systems.

\textbf{Limitation.}
\NoStateModels also remain some limitations.
First, vision-only policies might exhibit sensitivity to the background: changing the background (e.g., relocating the robot and table) may require additional fine-tuning to restore performance.
And in dual-arm settings, if only one arm is used for working, distribution shifts in the unused arm’s visual input may occasionally lead to unexpected movements of the unused arm.

% \textbf{Future Work.}
% In future work, we will further explore robot sensor designs for extreme spatial generalization and develop a dedicated simulation benchmark to standardize the evaluation of spatial generalization.

%% file: Sections/9_appendix.tex
\section{Appendix}
\subsection{Challenges With State Input}
\label{sec:app_state_challenge}

\input{Tables/state_challenge}

In the \textit{\say{Pick Pen}} task, we evaluate several strategies to improve height generalization of state-based policies with dual wide-angle \eefcams, including: (1) adding random noise ($[-5\ \text{cm},\ 5\ \text{cm}]$) to state heights, (2) collecting data at table heights 75–84 cm (30 episodes each 1 cm interval), (3) task-mixed training on \textit{\say{Pick Pen}} and \textit{\say{Pick Bottle}}, and (4) LoRA fine-tuning.
As reported in Table~\ref{tab:state_challenge}, none of these methods yields fundamental improvement, confirming that state inputs inherently limit spatial generalization.

\subsection{In-domain Performance}
\label{sec:app_indomain_exp}

% [Arxiv 版本添加的一段话]
In this section, we will report the in-domain performance of both \YesStateModels and \NoStateModels in our real-world tasks.
At the same time, we also report their performance on the LIBERO benchmark.

\input{Tables/in_domain}

In Table~\ref{tab:in_domain}, we report the in-domain success rates for different real-world tasks using policies with and without state input, with dual wide-angle \eefcams.
Even after removing the state input, the policies still maintain comparable performance on in-domain tasks, as the distribution of visual observations remains fully consistent with training.

\input{Tables/simulation}

In the simulation environment, we compare the in-domain performance of the $\pi_0$ policy with and without state input on the LIBERO benchmark.
As reported in Table~\ref{tab:simulation}, even in simulation environments with severely limited camera views, \NoStateModels achieve performance as perfect as \YesStateModels, and in some cases even surpass them, demonstrating the strong practicality of \NoStateModels.

%% file: Tables/state_challenge.tex
\begin{table}[htbp]
  \centering
  \caption{Height generalization performance of \YesStateModels under different optimization strategies.}
  \label{tab:state_challenge}
  \begin{tabular}{l c}
    \toprule
    Optimization strategy & Height Generalization \\
    \midrule
    \textbf{w/o state input} & \textbf{0.983} \\
    Random noise augmentation & 0.633 \\
    Diverse data collection & 0.117 \\
    Task-mixed fine-tuning & 0 \\
    LoRA fine-tuning & 0 \\
    \bottomrule
  \end{tabular}
\end{table}

%% file: Tables/in_domain.tex
\begin{table}[htbp]
    \centering
    \caption{In-domain success rates in different tasks using policies with and without state input.}
    \label{tab:in_domain}
    \begin{tabular}{@{}lcc@{}}
        \toprule
        Task name                & w/ state input & w/o state input \\ \midrule
        \textit{Pick Pen}                 & 1.0            & 1.0               \\
        \textit{Pick Bottle}              & 1.0            & 1.0               \\
        \textit{Put Lid}                  & 1.0            & 1.0               \\
        \textit{Fold Shirt}               & 1.0            & 0.967             \\
        \textit{Fetch Bottle (whole-body)} & 0.900          & 0.933               \\ \bottomrule
    \end{tabular}
\end{table}

%% file: Tables/simulation.tex
\begin{table}[htbp]
    \centering
    \caption{Simulation evaluations on the Libero benchmark, using policies with and without state input.}
    \label{tab:simulation}
    \begin{tabular}{lcc}
        \toprule
        Evaluation suite & w/ state input & w/o state input \\
        \midrule
        \textit{Libero Goal}    & 0.942 & 0.956 \\
        \textit{Libero Object}  & 0.964 & 0.962 \\
        \textit{Libero Spatial} & 0.968 & 0.976 \\
        \textit{Libero 10}      & 0.876 & 0.886 \\
        \midrule
        Average & 0.938 & 0.945 \\
        \bottomrule
    \end{tabular}
\end{table}